\documentclass[11pt]{article}

\usepackage[T1]{fontenc}
\usepackage[utf8]{inputenc}
\usepackage{lmodern}
\usepackage{graphicx}
\usepackage{tikz}
\usepackage{tabularx}
\usepackage{authblk}
\usepackage[a4paper,margin=1in]{geometry}
\usepackage{enumitem}
\usepackage{booktabs}
\usepackage[backend=biber,style=authoryear,maxcitenames=2]{biblatex}
\usepackage[hidelinks]{hyperref}
\usepackage{xurl}
\usepackage{comment}
\excludecomment{AIignore}

\title{What Counts as Strategic Reasoning? A Systematic Mapping of Chess Research on Humans, Engines, and Language Models}

\author[1]{Paolo Ciancarini%
\thanks{\href{https://orcid.org/0000-0002-7958-9924}
{ORCID: 0000-0002-7958-9924}}}

\author[2]{Remo Pareschi%
\thanks{Corresponding author: 
\href{mailto:remo.pareschi@unimol.it}{remo.pareschi@unimol.it}; 
\href{https://orcid.org/0000-0002-4912-582X}
{ORCID: 0000-0002-4912-582X}}}

\affil[1]{Department of Computer Science and Engineering, University of Bologna, Bologna, Italy}
\affil[2]{Stake Lab, University of Molise, Campobasso, Italy}

\date{}

\begin{document}
\maketitle

\begin{abstract}
Chess has long served as a model domain for studying search, expertise,
decision-making, and artificial intelligence. The emergence of large
language models (LLMs) has renewed the relevance of chess as a controlled
environment for investigating strategic reasoning and comparing human
and artificial decision-making. We present a systematic mapping study
of recent research spanning human players, classical chess engines,
neural and reinforcement-learning systems, LLMs, and hybrid approaches.
The final map comprises 84 core study families, classified according to
agent type, strategic-reasoning stages, and evaluation dimensions.

The map reveals a literature strongly concentrated on situation
assessment, evaluation, and action selection, while explicit planning,
explanation, metacognition, and human--AI collaboration remain less
explored. LLM research places particular emphasis on state
representation and generalization, whereas grounded explanation appears
more frequently in hybrid approaches combining language models with
engines, expert knowledge, or other external structures. Two distinctions 
emerge that the map aggregates rather than resolves: hybrid systems differ in 
where and when heterogeneous capabilities combine, and evaluations 
that show improved human performance do not thereby establish human--AI synergy. 
We propose both as extensions of the mapping framework.

We argue that
chess provides a useful bridge between cognitive and computational
perspectives on strategic reasoning, and identify explicit planning,
grounded and faithful explanation, metacognitive calibration, and
human--AI complementarity as directions for future research.
\end{abstract}


\section{Introduction}
\label{sec:introduction}

Chess became a scientific model of strategic reasoning through two
parallel research traditions. In cognitive science, de Groot's
pioneering studies of expert move choice investigated how chess players
perceive a position, generate candidate moves, analyze alternatives,
and reach a decision \parencite{degroot1965}. Chase and Simon later
emphasized the role of perceptual organization and pattern recognition
in chess expertise, showing that strong play cannot be understood
simply as the exploration of a larger search space
\parencite{chase_simon_1973}. In artificial intelligence, Shannon's
foundational formulation of computer chess approached the same problem
computationally, decomposing move choice into game-tree search and
position evaluation \parencite{shannon1950}. These traditions asked
closely related questions---how a decision maker represents a position,
explores alternatives, evaluates consequences, and selects an
action---but operationalized strategic reasoning in fundamentally
different ways.

The development of computer chess progressively changed the balance
between these traditions. 
Classical engines refined increasingly powerful search and evaluation techniques. Deep Blue's 1997 victory over Kasparov brought the approach to public attention \parencite{campbell2002deepblue}, but development continued well beyond it, and the strongest present-day engines, notably Stockfish, far exceed its playing strength.
Neural and
reinforcement-learning approaches introduced learned representations,
policies, and value functions, while systems such as AlphaZero retained
explicit search within a largely learned architecture
\parencite{silver2018general}. At the same time, research on human chess
continued to investigate expertise, intuition, selective search, time
allocation, and decision making. Chess thus developed into an unusual
research domain in which human and artificial approaches to the same
strategic decisions can be studied against a common formal environment.

Large language models (LLMs) introduce a further change. Unlike
conventional chess engines, they are not designed around an explicit
game-tree search architecture. They can process games and positions as
symbolic sequences, predict moves, answer questions about positions,
and generate natural-language explanations
\parencite{feng_chessgpt_2023,karvonen_emergent_2024}. Their use in
chess also introduces distinctive evaluation problems. A model must
maintain a coherent representation of the board and the rules; apparent
chess knowledge may reflect either transferable competence or patterns
encountered during training; and fluent explanations may be incorrect
or disconnected from the process that produced the selected move.

These developments make the meaning of \emph{strategic reasoning}
itself a central issue. Selecting a strong move is not necessarily
evidence of strategic planning. Representing a position correctly does
not imply understanding its strategic consequences, and producing a
plausible explanation does not establish that the explanation is
correct or faithful. Conversely, different research traditions may
study important components of strategic reasoning without using the
same terminology or evaluation criteria. Human studies investigate
constructs such as expertise, intuition, risk, and metacognition;
classical computer-chess research emphasizes search, evaluation, and
playing strength; human-aligned models may optimize similarity to human
decisions rather than maximal strength
\parencite{mcilroyyoung_aligning_2020,tang_maia2_2024}; and recent LLM
research increasingly examines state representation, generalization,
and explanation.

The contemporary literature is therefore both rich and fragmented.
Studies involving humans, classical engines, neural and
reinforcement-learning systems, LLMs, and hybrid approaches address
different parts of a common strategic-choice problem, but they often
use different concepts and different standards of evidence. As a
result, it is difficult to determine which components of strategic
reasoning have actually been investigated across agent types, how those
components are evaluated, and which combinations remain comparatively
underexplored.

To address this problem, we conduct a Systematic Mapping Study (SMS) of
recent research using chess to investigate strategic reasoning. Rather
than attempting to synthesize a common effect size across heterogeneous
research traditions, we map the literature along three multi-label
dimensions: \emph{agent type}, \emph{strategic-reasoning stage}, and
\emph{evaluation dimension}. The resulting matrices provide a common
analytical space in which human and artificial approaches can be
compared without assuming that they implement the same reasoning
process or pursue the same notion of successful performance.

The paper makes four contributions. First, it provides a systematic
cross-disciplinary map connecting contemporary research on human chess
cognition, classical computer chess, neural and reinforcement-learning
systems, LLMs, and hybrid approaches. Second, it proposes an operational
framework for distinguishing components of strategic reasoning from
the methods used to evaluate them. Third, it uses the resulting map to
identify underexplored intersections, with particular attention to
explicit planning, grounded and faithful explanation, metacognitive
calibration, and human--AI complementarity. Fourth, 
it identifies two distinctions compressed by the operational mapping—different structural 
and temporal forms of hybridization, 
and human augmentation versus genuine human--AI synergy—and proposes 
them as extensions of the framework.

Section~\ref{sec:strategic-traditions} reviews complementary conceptions of strategic reasoning in chess and provides the conceptual background for the mapping framework. Section~\ref{sec:research-questions} formulates the research questions, and Section~\ref{sec:method} describes the systematic mapping method, including the search, selection, and classification procedures. Section~\ref{sec:results} presents the resulting maps and identifies comparatively underexplored intersections. Section~\ref{sec:discussion} discusses their implications for strategic reasoning and human--AI systems. The paper then addresses threats to validity and concludes with directions for future research.

\section{Conceptions of Strategic Reasoning in Chess}
\label{sec:strategic-traditions}

Long before chess became a testbed for artificial intelligence, leading
players attempted to make explicit the principles and processes underlying
strategic decisions. Their accounts were primarily normative rather than
cognitive theories, since they sought to explain how a strong player
\emph{should} understand a position and organize action. Taken together,
however, they reveal a progressive enrichment of what strategic reasoning
in chess has been understood to involve.

A decisive step toward systematic positional reasoning was made by
Steinitz. In \emph{The Modern Chess Instructor}, he proposed that analysis
should be grounded in general ideas and in ``reasoning by analogies of
positions'', so that principles derived from one situation could support
judgement in structurally related ones \parencite{steinitz1889modern}.
Strategy consequently became more than the search for immediate tactical
opportunities: positional features could be evaluated and related to
general maxims capable of guiding subsequent action.

Nimzowitsch extended this project by giving positional strategy a more
explicit vocabulary. Concepts such as blockade, outposts, overprotection,
pawn-chain strategy, and prophylaxis made strategically relevant
relationships objects of conscious analysis
\parencite{nimzowitsch1929system}. His contribution was therefore not
simply a collection of principles, many of which had antecedents in
earlier practice, but their organization into knowledge that players
could deliberately use to interpret positions and construct plans.

Possessing such knowledge does not, however, determine how a decision
should be reached. Kotov addressed this complementary problem in
\emph{Think Like a Grandmaster}, where candidate moves and the ``tree of
analysis'' organize the generation and investigation of alternatives
\parencite{kotov1971think}. Strategic reasoning thereby acquires an
explicit procedural dimension in which positional judgement, calculation,
comparison, and eventual commitment perform distinguishable roles.

A parallel formulation of the problem of selecting what deserves analysis
appeared in the earliest computer-chess literature. Shannon distinguished
Type A strategies, based on relatively uniform search, from Type B
strategies that selectively investigate promising continuations
\parencite{shannon1950}. Since exhaustive analysis is generally
impractical, strategic choice becomes partly a problem of allocating
limited reasoning resources to relevant alternatives---a constraint
shared, although addressed differently, by human and computational
reasoners.

Chase and Simon add a cognitive perspective to this picture. Expert players do not overcome chess complexity simply by searching more extensively: chunking and pattern recognition allow meaningful configurations to be recognized rapidly, shaping which alternatives receive further analysis \parencite{chase_simon_1973}. Strategic reasoning therefore depends also on expertise-dependent representations of what is strategically salient.

These traditions reveal complementary aspects of strategic thought.
Principles, patterns, and heuristics constitute reusable
\emph{strategic knowledge}; reasoning determines their relevance,
generates and explores courses of action, reconciles competing
considerations, and supports commitment to a decision. Strategic
competence additionally requires control over when such knowledge and
processes should be deployed and when further analysis is warranted.

Computer chess made some of these functions operational and measurable,
while leaving open a broader question: which forms of strategic reasoning
are actually investigated when humans and different kinds of artificial
agents are studied? This question motivates the systematic mapping
developed in the remainder of the paper.

\section{Research Questions}
\label{sec:research-questions}

The parallel traditions outlined above suggest that strategic reasoning
in chess cannot be characterized along a single dimension. Human
cognition, classical computer chess, neural and reinforcement-learning
systems, LLMs, and hybrid approaches all investigate aspects of
strategic decision making, but differ in three fundamental respects:
\emph{who or what is taken as the reasoning agent}, \emph{which
components of the reasoning process are made explicit}, and \emph{what
evidence is accepted as demonstrating strategic competence}. The
purpose of the mapping is therefore not to rank agent classes or
aggregate their performance, but to characterize how these dimensions
are represented and related across the literature.

These considerations motivate four research questions. The first
characterizes the \emph{agent landscape}. The second examines how
strategic reasoning is decomposed and which components are actually
operationalized. The third investigates the \emph{evaluation practices}
used to establish strategic competence across agent types. Finally,
by combining these dimensions, the fourth shifts attention from
individual studies to the structure of the literature itself,
identifying which intersections are well represented and which remain
comparatively underexplored.

\begin{description}[style=nextline,leftmargin=2em]
  \item[\textbf{RQ1 --- Agent landscape.}]
  Which types of agents are studied in recent chess research on
  strategic reasoning?

  \item[\textbf{RQ2 --- Strategic-reasoning processes.}]
  Which stages of strategic reasoning are operationalized and
  evaluated across these agent types?

  \item[\textbf{RQ3 --- Evaluation practices.}]
  Which evaluation dimensions are used to assess strategic capabilities,
  and how do they vary across agent types?

  \item[\textbf{RQ4 --- Research gaps.}]
  Which combinations of agent types, strategic-reasoning stages, and
  evaluation dimensions remain comparatively underexplored?
\end{description}

Together, these questions define the two principal maps developed in
the study: \emph{Agent Type $\times$ Strategic-Reasoning Stage} and
\emph{Agent Type $\times$ Evaluation Dimension}. RQ1--RQ3 characterize
the populated regions of these maps, while RQ4 directs attention to
their sparse and empty regions.

\section{Method}
\label{sec:method}

\subsection{Study Design}

We conducted a Systematic Mapping Study (SMS) to characterize recent
research in which chess is used to investigate strategic reasoning across
human and artificial agents. An SMS was preferred to a systematic
literature review aimed at effect synthesis because the literature spans
heterogeneous research traditions, including cognitive studies of human
expertise, classical computer-chess research, neural and reinforcement
learning systems, large language models (LLMs), and hybrid human--AI or
multi-component systems. These traditions differ substantially in their
research questions, experimental designs, and evaluation criteria.

Accordingly, the objective of the study is not to estimate a common
effect size, but to identify and classify the main areas of research,
their relationships, and sparsely investigated combinations. The mapping
is organized along three dimensions: agent type, strategic-reasoning
stage, and evaluation dimension. The complete study protocol, search
strategies, codebook, and derived study-family dataset are provided in
the replication package.

\subsection{Scope and Information Sources}

The quantitative map covers work published from 2023 through 2026.
This interval was selected to capture the rapid emergence of
LLM-based approaches while retaining contemporary work on human
cognition, classical engines, and neural chess systems.

The search combined three sources: Scopus, the Web of Science Core
Collection, and a curated Zotero seed collection assembled from prior
work on computer chess, chess cognition, and AI. Scopus served as the
primary bibliographic database. Web of Science was added as a
complementary source, particularly to improve coverage of work on
human cognition, expertise, decision making, and planning. The Zotero
collection was used as a seed and cross-checking source rather than as
a substitute for the database searches.

The Scopus search was organized into five retained search families,
each designed to capture a different part of the interdisciplinary
research space:

\begin{table}[htbp]
\centering
\caption{Search families used in the systematic mapping study.}
\label{tab:search-families}
\begin{tabular}{@{}ll@{}}
\toprule
\textbf{Family} & \textbf{Primary focus} \\
\midrule
S1  & LLM reasoning and chess understanding \\
S2a & Explanation, commentary, and grounding \\
S2b & Human--AI interaction and collaboration \\
S3  & Generalization, robustness, and memorization \\
S5  & Human cognition, expertise, and decision making \\
WoS & Complementary coverage of cognition, planning, and expertise \\
\bottomrule
\end{tabular}
\end{table}

Two broader pilot searches were also explored during query development
but produced excessive noise and were subsequently refined rather than
retained as independent search families. This iterative search design
was intended to balance breadth with sufficient specificity across
research traditions whose terminology differs substantially.

The exact executable queries, execution dates, raw retrieval counts,
and search decisions are reported in the replication package. Keeping
the complete query strings outside the main text allows the search to
be reproduced without overloading the manuscript with database-specific
syntax.

\subsection{Record Consolidation and Study Families}

Records retrieved from the different searches were consolidated before
mapping. Duplicate detection used DOI and database identifiers where
available, followed by normalized-title comparison and manual inspection
of ambiguous cases.

The unit of analysis is a \emph{study family}, rather than an individual
bibliographic record. This distinction is important in the recent AI
literature because the same research contribution may appear first as
an arXiv preprint and subsequently as a conference paper, journal
article, or book chapter. Counting these manifestations independently
would inflate the apparent size of rapidly moving research areas.

When a preprint and a formally published version represented the same
study, they were therefore consolidated into one study family. The
published version was preferred when available; otherwise, the most
recent accessible version was used for coding. Distinct papers derived
from the same project were retained separately when they reported
substantively different research questions, methods, or evaluations.

\subsection{Eligibility and Classification}

Screening was conducted at the level of study families. The central
eligibility criterion was whether chess constituted a substantive
empirical or computational domain for investigating constructs relevant
to strategic reasoning.

Included studies were divided into two classes:

\begin{description}[style=nextline,leftmargin=2em]
    \item[\textbf{Core.}]
    Studies in which chess is a substantive empirical or model domain
    and which directly investigate at least one construct represented
    in the mapping framework, including human decision making or
    expertise, state representation, search, planning, evaluation,
    action selection, explanation, metacognition, human-likeness,
    generalization, or human--AI interaction.

    \item[\textbf{Context.}]
    Studies that contribute relevant conceptual, theoretical, review,
    transfer, or background evidence but do not directly instantiate
    the main chess-centered mapping constructs. This category also
    includes work in which chess is secondary to a broader experimental
    domain.
\end{description}

Context studies were retained to support interpretation and conceptual
framing but were excluded from the quantitative mapping matrices. The
final mapped corpus contains 84 Core study families and 19 Context
study families.

\subsection{Mapping Framework}

Each Core study was coded along three independent, multi-label
dimensions. Multi-label coding was necessary because a single study
may, for example, involve both a classical engine and an LLM, address
both position evaluation and move selection, and evaluate both move
quality and generalization.

The first dimension identifies the type of agent:

\begin{itemize}
    \item \textbf{H}: Human;
    \item \textbf{CE}: Classical/search engine;
    \item \textbf{NRL}: Neural or reinforcement-learning system;
    \item \textbf{LLM}: Large language model;
    \item \textbf{HYB}: Hybrid, centaur, or multi-component system.
\end{itemize}

The second dimension decomposes strategic reasoning into nine stages:

\begin{description}[style=nextline,leftmargin=2em]
    \item[\textbf{R1 --- State representation and rule grounding.}]
    Maintaining or inferring a coherent representation of the board,
    pieces, rules, and state transitions.

    \item[\textbf{R2 --- Situation assessment and pattern recognition.}]
    Identifying salient positional features, motifs, concepts, or
    strategic characteristics of the current position.

    \item[\textbf{R3 --- Candidate generation.}]
    Generating or prioritizing plausible actions or plans before deeper
    comparison.

    \item[\textbf{R4 --- Calculation, search, and look-ahead.}]
    Exploring concrete future continuations and their consequences.

    \item[\textbf{R5 --- Strategic planning and foresight.}]
    Constructing or comparing longer-horizon plans or goals that
    organize sequences of actions beyond immediate tactical calculation.

    \item[\textbf{R6 --- Evaluation and trade-off reasoning.}]
    Comparing alternatives in terms of positional value, risk,
    uncertainty, resources, or competing objectives.

    \item[\textbf{R7 --- Decision and action selection.}]
    Selecting a move or plan from the available alternatives.

    \item[\textbf{R8 --- Explanation and verbalization.}]
    Producing or evaluating a human-understandable account of why a
    position, move, or plan is good or bad.

    \item[\textbf{R9 --- Reflection, calibration, and metacognition.}]
    Assessing confidence, uncertainty, errors, reasoning quality, or
    the need to revise or continue the reasoning process.
\end{description}

The third dimension describes how the studied capability is evaluated:

\begin{itemize}
    \item \textbf{E1}: legality and state consistency;
    \item \textbf{E2}: move quality and engine agreement;
    \item \textbf{E3}: playing strength and game outcome;
    \item \textbf{E4}: strategic understanding;
    \item \textbf{E5}: explanation correctness or usefulness;
    \item \textbf{E6}: reasoning faithfulness;
    \item \textbf{E7}: human-likeness or skill calibration;
    \item \textbf{E8}: calibration, risk, or uncertainty;
    \item \textbf{E9}: generalization, memorization, or robustness;
    \item \textbf{E10}: human--AI synergy or complementarity.
\end{itemize}

The complete operational definitions are reported in the codebook
distributed with the replication package.

\subsection{Coding and Verification Procedure}

Coding was performed iteratively using the predefined multi-label
framework. Particular care was taken to distinguish terminology used
in a paper's motivation from constructs that were actually studied or
measured. For example, a paper describing its system as performing
``reasoning'' was not automatically assigned search or planning codes,
and high move quality alone was not treated as evidence of explicit
look-ahead or strategic planning.

Following the initial coding, the included study families underwent a
verification pass in which bibliographic information, eligibility
decisions, and assigned codes were checked against the available
full text or, where necessary, the strongest available source-level
evidence. During this audit, codes were retained only when the
corresponding construct was directly studied, operationalized, or
evaluated. Codes inferred solely from terminology, architectural
background, or broad motivational claims were removed.

The resulting dataset should therefore be interpreted as a
systematically audited mapping produced through conservative coding.
Independent duplicate coding was not performed across the complete
corpus; consequently, the study does not report an inter-rater
agreement statistic. The potential effect of coder judgment is
addressed explicitly as a threat to construct validity.

\subsection{Mapping and Analysis}

The quantitative analysis uses only the 84 Core study families.
Because all three coding dimensions are multi-label, counts are not
mutually exclusive and percentages within a row or column need not sum
to 100\%.

The primary analytical artifact is the
\emph{Agent Type $\times$ Strategic-Reasoning Stage} matrix. Each cell
reports the number of Core study families containing both the
corresponding agent code and reasoning-stage code. This matrix is used
to identify dominant reasoning profiles and sparse or empty
intersections.

A second
\emph{Agent Type $\times$ Evaluation Dimension} matrix characterizes
how the different research traditions establish evidence for strategic
capability. Marginal distributions and temporal trends over 2023--2026
are used as complementary analyses.

The interpretation focuses not only on densely populated cells but also
on systematically sparse combinations. Such cells identify candidate
research gaps, provided that their interpretation remains consistent
with the operational definitions in the codebook.

\section{Results}
\label{sec:results}

The quantitative analysis includes the 84 Core study families.
Because agent types, reasoning stages, and evaluation dimensions are
all multi-label, the counts reported below are not mutually exclusive.
A study may therefore contribute to several rows and columns of the
mapping matrices.

The results are organized around the two principal maps of the study:
the relationship between agent types and strategic-reasoning stages,
and the relationship between agent types and evaluation dimensions.
We then examine temporal trends and use the sparse intersections of
the maps to identify underexplored areas.

\subsection{Overview of the Research Landscape}
\label{sec:results-overview}

The 84 Core study families span all five agent categories. Human-centered
studies constitute the largest category (38/84, 45.2\%), followed by
Classical/search Engine studies (37/84, 44.0\%), LLM studies
(28/84, 33.3\%), Neural/RL studies (24/84, 28.6\%), and Hybrid studies
(19/84, 22.6\%). These percentages do not sum to 100\% because a study
may involve more than one type of agent.

At the level of strategic reasoning, the literature is dominated by
R7 Decision/action selection (63/84, 75.0\%), followed by R2 Situation
assessment (43/84, 51.2\%) and R6 Evaluation/trade-offs
(40/84, 47.6\%). R1 State representation appears in 30 studies and
R4 Calculation/search in 24. R5 Strategic planning is represented in
17 studies and R3 Candidate generation in 12. The least populated
stages are R8 Explanation/verbalization (9 studies) and
R9 Reflection/metacognition (10 studies).

The marginal distributions provide an initial picture of the field,
but they obscure substantial differences between research traditions.
These differences become visible in the two mapping matrices.

\subsection{Agent Types and Strategic-Reasoning Stages}
\label{sec:agent-reasoning}

Table~\ref{tab:agent-reasoning} maps agent types to the nine
strategic-reasoning stages. Each cell reports the number of Core study
families containing both the corresponding agent code and reasoning
code. The values should therefore be interpreted as intersections in a
multi-label map, rather than as mutually exclusive classifications.

\begin{table*}[htbp]
\centering
\caption{Mapping of agent types to strategic-reasoning stages.
Cell values are numbers of Core study families. Both dimensions are
multi-label; row totals therefore do not equal the number of studies
in the corresponding agent category.}
\label{tab:agent-reasoning}
\small
\setlength{\tabcolsep}{4pt}
\begin{tabular}{@{}lrrrrrrrrr@{}}
\toprule
\textbf{Agent} &
\textbf{R1} &
\textbf{R2} &
\textbf{R3} &
\textbf{R4} &
\textbf{R5} &
\textbf{R6} &
\textbf{R7} &
\textbf{R8} &
\textbf{R9} \\
\midrule
Human (H), $n=38$             & 4  & 17 & 6 & 6  & 3 & 14 & 31 & 0 & 8 \\
Classical Engine (CE), $n=37$ & 9  & 12 & 4 & 13 & 7 & 26 & 30 & 6 & 5 \\
Neural/RL (NRL), $n=24$       & 10 & 12 & 5 & 8  & 6 & 14 & 18 & 2 & 0 \\
LLM, $n=28$                   & 20 & 17 & 4 & 10 & 8 & 12 & 19 & 6 & 3 \\
Hybrid (HYB), $n=19$          & 3  & 10 & 5 & 5  & 3 & 13 & 14 & 7 & 5 \\
\bottomrule
\end{tabular}
\end{table*}

The first pattern is the prominence of \emph{decision/action selection}.
R7 is the most populated stage for Human, Neural/RL, and Hybrid studies,
and is also highly represented in Classical Engine and LLM research.
It occurs in 31 of the 38 Human studies, 30 of the 37 Classical Engine
studies, 18 of the 24 Neural/RL studies, 19 of the 28 LLM studies, and
14 of the 19 Hybrid studies. Across otherwise different research
traditions, selecting an action is therefore the most consistently
studied component of strategic reasoning.

The profiles preceding action selection, however, differ substantially.
Classical Engine studies display the clearest
\emph{search--evaluation--decision} profile: R4 Calculation/search
appears in 13 studies, R6 Evaluation in 26, and R7 Decision in 30.
This reflects the traditional computer-chess architecture in which
explicit search and position evaluation jointly support move selection.

Neural/RL studies retain a strong decision-oriented profile but place
comparatively greater emphasis on learned representations and
assessment. R1 and R2 each occur in substantial parts of this
literature, whereas explanation and reflection are almost absent.
Only two Neural/RL studies address R8 and none is coded for R9.

The LLM profile is distinctive. R1 State representation appears in
20 of 28 LLM studies, making it particularly prominent in this
category. R2 Situation assessment appears in 17 and R7 Decision in 19.
R4 Calculation/search is present in 10 studies and R5 Strategic planning
in eight. These counts show that contemporary LLM chess research is not
limited to final move prediction: substantial attention is devoted to
whether models can maintain coherent board representations, assess
positions, and perform forms of look-ahead. Nevertheless, explicit
candidate generation, explanation, and reflection remain comparatively
weakly represented.

Human studies exhibit a different pattern. They concentrate on
R7 Decision (31), R2 Assessment (17), and R6 Evaluation (14), while
R9 Reflection/metacognition appears in eight studies, proportionally
more than in any other agent category. By contrast, only three Human
studies directly operationalize R5 Strategic planning.

The empty Human $\times$ R8 cell requires careful interpretation.
It does not imply that human chess players do not verbalize their
reasoning, nor that verbal protocols are absent from chess research.
Rather, under the operational definition adopted in this mapping, none
of the recent Core studies makes the quality or properties of human
strategic explanation itself a primary evaluated construct.

Hybrid systems provide the strongest contrast. Seven of the 19 Hybrid
studies address R8 Explanation, a substantially larger proportion than
in the other agent classes. The explanation-oriented systems in this
group commonly introduce external structure---for example, engine
evaluations, expert models, symbolic rules, concept representations,
knowledge graphs, or strategic taxonomies---to ground the generated
explanation. The pattern therefore suggests, without establishing a
causal claim, that grounded chess explanation is currently more
frequently investigated in hybrid architectures than in standalone
agent paradigms.

Finally, the matrix exposes a methodological distinction between
decision performance and evidence of a reasoning process. R7 is far
more common than R3 Candidate generation or R5 Strategic planning.
Consequently, successful move selection should not by itself be
interpreted as evidence that candidate generation, look-ahead, or
longer-horizon planning has been demonstrated.

\subsection{Agent Types and Evaluation Dimensions}
\label{sec:agent-evaluation}

The second map concerns the evidence used to evaluate strategic
capability. Table~\ref{tab:agent-evaluation} reports the intersections
between agent types and the ten evaluation dimensions.

\begin{table*}[htbp]
\centering
\caption{Mapping of agent types to evaluation dimensions.
Cell values are numbers of Core study families. Agent and evaluation
codes are multi-label.}
\label{tab:agent-evaluation}
\small
\setlength{\tabcolsep}{3.5pt}
\begin{tabular}{@{}lrrrrrrrrrr@{}}
\toprule
\textbf{Agent} &
\textbf{E1} &
\textbf{E2} &
\textbf{E3} &
\textbf{E4} &
\textbf{E5} &
\textbf{E6} &
\textbf{E7} &
\textbf{E8} &
\textbf{E9} &
\textbf{E10} \\
\midrule
Human (H), $n=38$             & 1  & 14 & 7  & 12 & 0 & 0 & 9 & 8 & 1  & 5 \\
Classical Engine (CE), $n=37$ & 4  & 23 & 15 & 7  & 4 & 2 & 9 & 6 & 6  & 6 \\
Neural/RL (NRL), $n=24$       & 1  & 8  & 12 & 5  & 2 & 2 & 9 & 0 & 7  & 2 \\
LLM, $n=28$                   & 18 & 16 & 8  & 11 & 5 & 4 & 2 & 2 & 13 & 0 \\
Hybrid (HYB), $n=19$          & 1  & 11 & 7  & 5  & 5 & 3 & 7 & 3 & 2  & 7 \\
\bottomrule
\end{tabular}
\end{table*}

The evaluation matrix shows that the research traditions differ not
only in what aspects of reasoning they investigate, but also in what
they accept as evidence of strategic capability.

Classical Engine research is dominated by E2 Move quality/engine
agreement and E3 Playing strength. E2 occurs in 23 of 37 Classical
Engine studies and E3 in 15. This is consistent with the traditional
computer-chess paradigm, in which capability is established through
position scores, tactical tests, game outcomes, or comparison with
strong reference engines.

Human studies show a more heterogeneous evaluation profile. E2 Move
quality remains prominent, often because engine evaluation provides an
external reference for human decisions, but E4 Strategic understanding,
E7 Human-likeness/skill calibration, and E8 Calibration/risk/uncertainty
also occur frequently. These dimensions reflect the cognitive and
behavioral emphasis on expertise, intuition, decision time, confidence,
risk, and differences between levels of human skill.

Neural/RL studies combine competitive and behavioral evaluation.
E3 Playing strength appears in 12 of 24 studies, while E7
Human-likeness appears in nine. This dual profile reflects two partly
different objectives in neural chess research: producing increasingly
strong agents and modeling human decision behavior.

LLM studies exhibit the most distinctive evaluation profile.
E1 Legality/state consistency occurs in 18 of 28 studies, substantially
more often than in any other agent class. E9
Generalization/memorization/robustness occurs in 13, and E2 Move quality
in 16. These concentrations reflect two central questions in recent LLM
chess research: whether a model can maintain a coherent representation
of the game state and rules, and whether apparent chess competence
transfers beyond familiar or potentially memorized distributions.

The importance of E9 is particularly relevant to recent benchmark
design. Evaluations based on novel positions, Chess960, structurally
modified games, or chess variants attempt to distinguish generalizable
strategic competence from retrieval of patterns represented in training
data. This direction complements conventional move-quality evaluation
by asking not only whether a model selects a strong action, but whether
the underlying competence survives changes in the task distribution.

Explanation-oriented evaluation remains sparse. E5 Explanation
correctness/usefulness appears in only six Core studies and E6 Reasoning
faithfulness in five. These dimensions are concentrated primarily in
LLM and Hybrid research. Their low frequency is significant because a
fluent natural-language explanation is not necessarily evidence that
the explanation is correct, grounded in the position, or faithful to
the process that produced the decision.

The Hybrid profile again differs from the other categories.
E10 Human--AI synergy appears in seven of 19 Hybrid studies, and E5
Explanation quality in five. Hybrid research therefore contains a
disproportionate share of work concerned with whether artificial
systems are not merely strong, but useful, understandable, and
complementary to human decision makers.

Conversely, no LLM study in the current Core corpus is coded for direct
evaluation of E10 Human--AI synergy. This absence is noteworthy given
the growing practical use of language models as interactive assistants.
It indicates a difference between deploying LLMs in assistance-oriented
settings and experimentally evaluating whether human--LLM teams
actually outperform, complement, or appropriately rely on their
individual components.

\begin{AIignore}
\subsection{Temporal Evolution of the Map}
\label{sec:temporal}

The temporal window we used captures a period in which LLM research enters an
already established landscape of human cognition, classical engines,
and neural chess systems. Figure~\ref{fig:temporal-agents} summarizes
the number of Core study families associated with each agent category
by publication year.

%

The temporal pattern should be interpreted as expansion rather than
simple paradigm replacement. Human-centered and engine-based research
remain active throughout the period, while LLM studies become an
increasingly visible component of the map. The emergence of LLM work
also changes the distribution of research questions: state tracking,
rule grounding, contamination, memorization, and transfer become more
prominent evaluation concerns.

This shift is particularly visible in the combination of R1 and E1/E9.
Earlier chess-AI paradigms could generally assume a stable symbolic game
state and focus evaluation on search quality or playing strength. For
LLMs, maintaining the state itself becomes an empirical question, while
the abundance of chess material in training corpora makes
generalization and contamination explicit methodological concerns.

The temporal analysis therefore suggests that the LLM wave has added
new evaluation problems to chess research rather than displacing its
older questions. Search, expertise, evaluation, and decision quality
remain central, but they now coexist with state consistency,
memorization, language-mediated explanation, and model robustness.

\end{AIignore}

\subsection{Open Spaces in the Map}
\label{sec:open-spaces}

The purpose of a systematic map is not only to identify concentrations
of research but also to reveal combinations that remain weakly
explored. Four open spaces are particularly visible across
Tables~\ref{tab:agent-reasoning} and
\ref{tab:agent-evaluation}.

\paragraph{The planning gap.}
R5 Strategic planning appears in only 17 Core study families, compared with 63 for R7 Decision/action selection. The difference matters: many studies show that an agent can choose a good move without directly testing whether it constructs or maintains a longer-horizon plan. The gap is especially relevant to claims about LLM reasoning, where successful action selection can result from several mechanisms that do not necessarily imply explicit planning. Few studies test the question directly, either through architectural constraints or by inspecting internal computation; the map does not currently distinguish these from studies that leave planning untested. 

\paragraph{The explanation and faithfulness gap.}
R8 Explanation is represented in nine studies, E5 Explanation quality
in six, and E6 Reasoning faithfulness in five. The map therefore
distinguishes three questions that are sometimes conflated: whether a
system can produce an explanation, whether that explanation is
strategically correct or useful, and whether it faithfully reflects
the reasoning that produced the decision. Current research addresses
the first more frequently than the latter two.

\paragraph{The metacognitive gap.}
R9 Reflection/metacognition occurs in only ten Core studies and is
concentrated disproportionately in human and hybrid research. Human
studies provide constructs such as confidence, stopping decisions,
thinking-time allocation, risk sensitivity, and reliance on advice,
whereas analogous forms of metacognitive control remain weakly
operationalized for artificial chess agents.

\paragraph{The collaboration gap.}
E10 Human--AI synergy occurs in only seven Core studies. In particular,
no LLM-only study in the current map directly evaluates human--AI
synergy. This is notable because contemporary applications increasingly
position AI as a trainer, analyst, commentator, or interactive
assistant rather than solely as an autonomous chess player. Direct
measurement of complementarity, reliance, delegation, and team
performance therefore represents a substantial open area.

\subsection{Synthesis}
\label{sec:results-synthesis}

Taken together, the two matrices reveal a field composed of distinct
but increasingly connected research traditions.

Human studies emphasize assessment, decision behavior, expertise, and
metacognition. Classical engines exhibit a search--evaluation--decision
profile and rely heavily on move quality and playing strength as
evidence. Neural/RL systems combine learned representations and action
selection with competitive and human-behavioral evaluation. LLM studies
place unusually strong emphasis on state representation, legality,
assessment, action selection, and generalization. Hybrid systems
disproportionately address explanation and human--AI complementarity.

The resulting map also reveals a broader transition in how strategic
AI can be evaluated. Traditional chess AI primarily asks whether an
agent can select strong moves. The emerging research agenda increasingly
asks whether an agent can maintain the relevant state, form and evaluate
plans, transfer its competence to unfamiliar settings, explain its
decisions correctly and faithfully, recognize uncertainty, and
collaborate effectively with humans.

These patterns provide the empirical basis for the gaps and implications
discussed in the following section.

\section{Discussion}
\label{sec:discussion}
The mapping shows that chess research does not rely on a single notion of strategic reasoning. Human cognition, classical computer chess, neural and reinforcement-learning systems, LLMs, and hybrid approaches emphasize different components of the strategic process and use different criteria to establish competence. The map's value lies in making these traditions comparable without assuming they implement the same reasoning mechanisms.

The discussion proceeds in two registers. One draws out what the map shows: that decision quality should not be equated with evidence of a reasoning process (Section~\ref{sec:discussion-decision}); that LLMs make state representation and generalization central evaluation problems (\ref{sec:discussion-llm}); that grounded explanation is increasingly investigated through hybrid architectures (\ref{sec:discussion-explanation}); and that metacognition remains comparatively underexplored (\ref{sec:discussion-metacognition}).

In parallel, we take up two distinctions that the map, in its present form, aggregates rather than resolves. The HYB category aggregates architectures that differ in where and when heterogeneous capabilities combine (\ref{sec:discussion-hybridization}), and the evaluation dimensions do not separate improvements in human performance from genuine human--AI synergy (\ref{sec:discussion-synergy}). Both are offered as interpretive instruments for reading the hybrid literature and as candidate extensions of the mapping framework.

Sections~\ref{sec:discussion-bridge} and~\ref{sec:discussion-implications} then consider chess as a bridge domain and the implications for future evaluation.

\subsection{Decision Quality Is Not Equivalent to Strategic Reasoning}
\label{sec:discussion-decision}

The dominance of action selection over candidate generation and
explicit planning highlights an important methodological distinction.
Chess provides unusually precise measures of decision quality: engine
evaluations, tactical test suites, and game outcomes can establish
whether a move is strong without revealing how that move was produced.
A convenient outcome measure can therefore become a proxy for the much
richer construct of strategic reasoning.

This distinction is relatively transparent in classical computer chess,
where search and evaluation are explicit architectural components. It
is less straightforward for learned systems. A neural policy or
language model may select a strong move without demonstrating that it
generated alternatives, searched future states, or constructed a
longer-horizon plan. Conversely, weak full-game performance does not
imply the absence of every relevant reasoning capability: a system may
assess positions or identify strong individual moves while failing to
sustain coherent decisions over an entire game.

Recent work has begun to address the gap from the opposite direction, by constraining or directly inspecting the computation rather than inferring process from behaviour. \textcite{ruoss2024amortized} train
transformers on engine-annotated positions and reach grandmaster-level
play with no explicit search, which establishes that strong action
selection can be produced without a search procedure at inference
time. \textcite{jenner2024evidence} report the complementary result for
Leela Chess Zero: activations associated with future moves are causally
important for the current output, attention heads carry information
between the squares of earlier and later moves, and a probe recovers
the optimal move two turns ahead. Neither line of work settles what
these systems do in general, but both illustrate a form of evidence
that behavioural measures cannot supply, since the claim concerns the
computation rather than its output.

Chess-based evaluations should therefore distinguish \emph{decision
performance} from \emph{process evidence}, and, within the latter,
behavioural from internal evidence. Claims about strategic reasoning
become stronger when action quality is complemented by evidence
concerning representation, candidate generation, look-ahead, plan
formation, or revision. The R1--R9 framework provides a vocabulary for
identifying which component is at issue; the evaluation dimensions
E1--E10, being defined over observable performance, do not currently
distinguish the kind of evidence brought to bear on it. E6 Reasoning faithfulness 
comes closest, but asks whether a stated explanation matches the decision process, 
not whether the process has been inspected directly.

\subsection{LLMs Shift the Problem Toward Representation and Generalization}
\label{sec:discussion-llm}

The LLM literature changes what must be demonstrated before strategic
competence can be assessed. In conventional chess software, the board
state and legal rules are represented explicitly by the program.
Search and evaluation operate on a state whose formal consistency is
largely guaranteed by construction. For a language model processing
moves or serialized positions as tokens, this assumption no longer
holds. Maintaining piece locations, legal transitions, and other state
variables becomes an empirical capability.

State tracking is therefore not merely a technical prerequisite. A
system cannot reason strategically about a position it represents
incorrectly. This helps explain why representation and state
consistency occupy such a prominent place in contemporary LLM chess
research.

A second problem concerns the distinction between transferable
strategic competence and exposure to chess material during training.
Standard chess is unusually vulnerable to this ambiguity because
opening theory, annotated games, tactical positions, instructional
texts, and online discussions are abundant in public corpora.
Consequently, recent evaluations increasingly use novel positions,
distribution shifts, Chess960, controlled perturbations, or modified
chess environments to test whether apparent competence survives changes
that weaken memorization as an explanation.

This development broadens chess's role as an AI benchmark.
Systematic modifications of rules, initial states, information, or
board structure can preserve important aspects of adversarial decision
making while testing whether strategic capabilities transfer beyond
familiar configurations.

\subsection{Hybridization Operates at Different Levels}
\label{sec:discussion-hybridization}

The operational HYB category used in the mapping deliberately groups together hybrid, centaur, and multi-component systems. This aggregation helps identify studies that combine otherwise distinct capabilities, but the resulting systems should not be treated as theoretically equivalent. We can distinguish at least three levels of hybridization, of which only the latter two necessarily involve interaction among distinct agents.

First, \emph{computational hybridization} occurs within an artificial system when different computational paradigms contribute complementary functions. AlphaZero provides a canonical example: learned policy and value representations are combined with explicit Monte Carlo Tree Search, so that learned pattern-based evaluation and forward search participate in the same decision architecture \parencite{silver2018general}. Such a system is hybrid at the level of computational mechanisms even though no human component is involved.

Second, \emph{multi-agent coordination} combines distinct human or artificial agents that preserve identifiable boundaries, roles, and a degree of decision autonomy while contributing to a collective process. A system-theoretical account of contemporary human--AI interaction explicitly distinguishes this autonomy-preserving paradigm from Centaurian integration: multi-agent systems coordinate heterogeneous agents through structured interaction while maintaining their individual identities \parencite{borghoff2025humanartificial}. Organizational models involving human supervisors, LLM-based consultants, and specialized AI workers provide a concrete instance of this architecture \parencite{borghoff2025organizational}.

Third, \emph{Centaurian integration} involves a deeper form of human--AI combination. Here the defining property is not simply cooperation among heterogeneous agents, but increasing functional interdependence within a joint decision process. In system-theoretical terms, whereas multi-agent systems resemble ecosystems of autonomous entities, Centaurian systems seek tighter integration analogous to symbiotic or organism-like configurations, with more permeable functional boundaries and a more unified locus of decision-making \parencite{borghoff2025humanartificial}. Questions of task allocation, decision authority, responsibility, and complementarity therefore become constitutive features of the architecture rather than merely external properties of its use. 

Chess supplies instances of the second and third levels. Kasparov introduced advanced chess in 1998, a single player using engines as an extension of their own analysis; the freestyle format, in which teams orchestrate several engines and human contributors, emerged separately and is described in his retrospective account \parencite{kasparov2010chessmaster}. The first is Centaurian in design, the second multi-agent coordination. Whether advanced chess achieved the functional interdependence its design intended, or settled into computer-assisted play with the human still deciding alone, is an empirical question rather than a definitional one, and Section~\ref{sec:discussion-synergy} returns to it.

The three distinctions above concern the \emph{structural locus} at which heterogeneous capabilities or components are combined. They are also predominantly synchronic, in that the combined capabilities contribute to the same decision architecture or decision episode.

A cross-cutting distinction therefore concerns the \emph{temporal coupling} between the contributions. Hybridization may also be \emph{diachronic}: an artificial system contributes not to the decision itself but to an agent's competence, which then decides without the artificial component in the loop. \textcite{schut2025bridging} provide a particularly clear chess example. They extract concept vectors from AlphaZero's internal representations, filter candidate concepts for novelty and teachability, and subsequently present selected concept prototypes to four elite grandmasters in a controlled learning protocol. The grandmasters' improved performance after the learning phase provides evidence that knowledge from a superhuman artificial system can transfer to human experts.

What makes this analytically relevant as diachronic hybridization rather than ordinary instruction is the provenance of the transferred knowledge. 
The concepts are selected because they contain information not represented in the human-play reference space, so the subsequently acquired competence is at least partly machine-derived rather than recovered from the human-game corpus. The artificial system contributes causally to later human decision-making, but no persistent joint decision architecture is required once the transfer has occurred. We call this configuration \emph{diachronic coupling}: artificial and human capabilities combine across successive phases of learning and decision rather than simultaneously at the point of action. 
\begin{figure}[t]
\centering
\begin{tikzpicture}[
  font=\small,
  cell/.style={draw, rounded corners=2pt, align=left,
               text width=4.5cm, inner sep=5pt, minimum height=1.5cm},
  hdr/.style={align=center, font=\small\bfseries},
  row/.style={anchor=east, align=right, font=\small\bfseries}
]
\node[hdr] at (2.6,1.15) {Synchronic};
\node[hdr] at (7.65,1.15) {Diachronic};

\node[row] at (-0.05,0)  {Computational};
\node[row] at (-0.05,-2) {Multi-agent};
\node[row] at (-0.05,-4) {Centaurian};

\node[cell] at (2.6,0)   {AlphaZero: learned evaluation and explicit MCTS in one decision architecture};
\node[cell] at (7.65,0)  {Engine-supervised network that then decides without explicit search};
\node[cell] at (2.6,-2)  {Freestyle and consultation chess: several human and engine agents deliberating, with the human arbitrating};
\node[cell] at (7.65,-2) {Machine-derived concepts taught to grandmasters who then play unaided};
\node[cell] at (2.6,-4)  {Advanced chess in Kasparov's sense: engines used as an extension of a single player's analysis};
\node[cell, densely dashed] at (7.65,-4) {\emph{Vacant by construction}};

\draw[->, thick] (0.175,1.6) -- (10.075,1.6);
\node[font=\footnotesize] at (5.125,1.9) {temporal coupling};

\draw[->, thick] (-2.9,0.85) -- (-2.9,-4.85);
\node[font=\footnotesize, rotate=90, anchor=south] at (-3.2,-2) {structural locus};
\end{tikzpicture}
\caption{Two cross-cutting dimensions of hybridization, with chess examples.
The structural locus concerns \emph{where} heterogeneous capabilities are
combined; the temporal coupling concerns \emph{when}.}
\label{fig:hybridization-dimensions}
\end{figure}
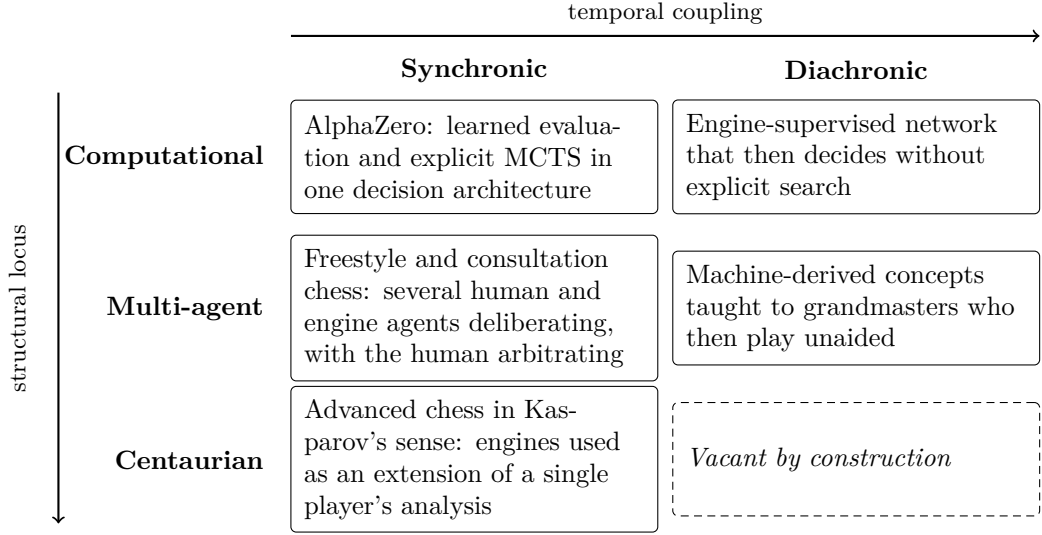

Figure~\ref{fig:hybridization-dimensions} shows the two dimensions together.
The Centaurian cell of the diachronic column is vacant by construction, not merely unpopulated. The vacancy follows from the definition rather than from the state of the literature: a coupling defined by functional interdependence within a shared decision episode cannot persist once that episode has dissolved. The combination therefore marks a boundary of the taxonomy rather than a research gap of the kind identified in Section~\ref{sec:open-spaces}.

The case also illustrates an evaluative asymmetry discussed in Section~\ref{sec:discussion-synergy}. Because the outcome of interest is improvement in human knowledge or performance, augmentation can be measured directly by comparing human performance before and after AI-mediated learning. By contrast, an artificial-agent baseline is not naturally defined on the same learning outcome, so the same comparison cannot establish the stronger criterion of human--AI synergy. Diachronic coupling is thus typically evaluated in augmentation terms, which reinforces the need to keep the architecture of human--AI interaction distinct from the performance relation used to evaluate it.

These distinctions caution against interpreting the HYB row of the mapping as representing a single architectural or cognitive paradigm. Its value is primarily operational: it captures studies that combine heterogeneous capabilities. Theoretical interpretation requires asking \emph{what} is being combined, \emph{at which structural level}, \emph{whether the coupling is synchronic or diachronic}, whether the components preserve independent agency, and whether human involvement amounts to coordination with artificial agents, deeper Centaurian integration, or knowledge transfer across successive phases of interaction. We offer this sub-classification as an interpretive instrument for reading the hybrid literature rather than as a coded dimension of the present map; applying it as a formal sub-dimension of the HYB category is a natural extension of the mapping framework.

\subsection{Grounded Explanation Is Emerging as a Hybrid Capability}
\label{sec:discussion-explanation}

The map distinguishes three questions that should not be conflated:
whether a system can produce an explanation, whether that explanation
is strategically correct and useful, and whether it is faithful to the
process or evidence supporting the decision. This distinction is
particularly important for LLMs, because linguistic fluency can create
a strong impression of understanding even when the underlying chess
analysis is incorrect or weakly grounded.

Recent explanation-oriented systems frequently address this problem by
introducing external structure. Engines can supply evaluations and
concrete variations; expert models can identify strategic concepts;
symbolic components can represent rules and constraints; and structured
knowledge can provide relations that an LLM can transform into
human-readable language. The concentration of explanation-oriented work
among hybrid approaches is therefore suggestive of an emerging design
pattern, although the current evidence is too limited to conclude that
hybridization is necessary for grounded explanation.

A useful distinction is consequently between \emph{linguistic
explanation} and \emph{grounded strategic explanation}. The latter
requires a relationship between the verbal account and independently
supported properties of the position, candidate moves, variations, or
plans. Chess is particularly suitable for studying this distinction
because many claims about material, threats, tactical sequences, and
move consequences can be independently checked.

\subsection{Metacognition Remains Underexplored}
\label{sec:discussion-metacognition}

Human chess research provides several ways of studying the decision
maker's relationship to its own reasoning: confidence, thinking time,
stopping decisions, risk sensitivity, error persistence, and reliance
on advice. These constructs concern not only which decision is made,
but whether additional reasoning is worthwhile, whether an evaluation
is reliable, and whether a previous conclusion should be reconsidered.

Comparable questions remain less developed for artificial chess agents.
Yet they become increasingly important when AI systems are intended to
assist rather than replace human decision makers. A system that produces
strong recommendations but cannot identify the conditions under which
its analysis is unreliable may be less useful than one whose confidence
is well calibrated.

Chess offers natural tests for machine metacognition. Confidence can be
compared with objective decision quality; additional computation can be
related to improvement in the resulting move; and revision can be
studied after feedback or contradictory evidence. For LLMs, such
experiments could also distinguish verbal expressions of uncertainty
from genuine calibration.





\subsection{From Human Augmentation to Dynamic Complementarity}
\label{sec:discussion-synergy}

The limited attention paid to human--AI synergy is particularly
significant because assistance, augmentation, and complementarity
should not be treated as equivalent. A human assisted by an artificial
system may perform better than the unaided human without the resulting
human--AI configuration outperforming the artificial system alone.
Recent work therefore distinguishes \emph{human augmentation} from
\emph{human--AI synergy} \parencite{vaccaro2024combinations}.

Let $P(X)$ denote performance on a given outcome measure for a
configuration $X$, where $H$ is the human component, $A$ the artificial
component, and $H+A$ their combination. Human augmentation can be
expressed as the gain

\[
\Delta_{\mathrm{aug}}
=
P(H+A)-P(H),
\]

so that $\Delta_{\mathrm{aug}}>0$ indicates that AI assistance improves
human performance. Genuine human--AI synergy requires the stronger
condition

\[
\Delta_{\mathrm{syn}}
=
P(H+A)-\max\{P(H),P(A)\},
\]

with $\Delta_{\mathrm{syn}}>0$ indicating positive synergy. Conversely,
$\Delta_{\mathrm{syn}}<0$ identifies a form of negative synergy, in
which the combined system performs worse than its stronger constituent.
The distinction makes explicit that positive augmentation does not
imply positive synergy: $P(H+A)$ may exceed $P(H)$ while remaining
below $P(A)$. This pattern is important empirically. A recent
meta-analysis across human--AI experiments found positive human
augmentation on average, but no corresponding average advantage of the
combination over the better individual component
\parencite{vaccaro2024combinations}.

The distinction also clarifies what different forms of evaluation can
establish. Where the same performance outcome can meaningfully be
measured for humans, artificial agents, and human--AI teams, all three
baselines are needed to demonstrate synergy. Where no meaningful
AI-alone baseline exists---for example, when the outcome concerns human
learning---improvement relative to the unaided human may establish
augmentation, but should not by itself be described as evidence of
synergy. Related work on Centaurian design has similarly emphasized
that combining human and artificial components does not by itself
guarantee superior joint performance
\parencite{pareschi2024beyond}.

Chess provides an unusually informative longitudinal setting in which
the distinction can be observed. Early Advanced Chess was motivated by
a plausible division of cognitive labour: engines could provide
tactical calculation and verification, while human players contributed
strategic judgement, candidate generation, and longer-horizon planning.
Kasparov's retrospective account of Advanced Chess and subsequent
freestyle play emphasized that the interaction process itself could
become a source of performance advantage: effective orchestration of
several engines could compensate for weaker individual chess ability
\parencite{kasparov2010chessmaster}.

As chess engines became substantially stronger, however, the basis of
this complementarity changed. Evidence from conventional, centaur, and
engine chess tournaments suggests that AI adoption produces both
substitution and complementation. Traditional chess-playing capability
becomes less predictive of performance once sufficiently strong engines
enter the decision process, while new human capabilities associated
with selecting, configuring, tuning, and governing artificial systems
become consequential \parencite{krakowski2023artificial}. Human
contribution therefore need not simply disappear as machine competence
increases; rather, the locus at which it adds value can migrate.

We refer to this changing relation as a \emph{dynamic complementarity
boundary}: the boundary between tasks or decision functions for which
human--AI integration yields positive synergy and those for which the
stronger component performs at least as well alone. Its position is not
fixed. It changes with the relative competence of human and artificial
agents, the allocation of subtasks, the design of the interaction
process, and the distribution of decision authority. A human
contribution that is complementary at one technological stage may
become redundant or detrimental at another, while new forms of
complementarity may simultaneously emerge at a different level of the
system.

This perspective also suggests an asymmetry when artificial competence
substantially exceeds human competence on a particular object-level
decision. Human intervention then has progressively fewer opportunities
to improve an already superior artificial choice while retaining
opportunities to degrade it. In this regime, human intervention can
function as an \emph{asymmetric error channel} at the object level even
while remaining valuable at higher levels such as objective setting,
tool selection, interpretation, orchestration, exception handling, or
governance.

The historical evolution of centaur chess should consequently not be
interpreted simply as either the success or the failure of human--AI
collaboration. It is better understood as a transition between
different regimes of complementarity. What initially appeared as
direct cooperation between human strategic judgement and machine
calculation can evolve toward human orchestration of increasingly
autonomous artificial capabilities. Chess thus provides a controlled
setting in which substitution, augmentation, genuine synergy, and
negative synergy can be distinguished rather than subsumed under a
single notion of ``hybrid'' performance.

This interpretation has direct implications for E10. Studies concerned
with human--AI complementarity should, wherever the outcome permits,
compare human performance, artificial-agent performance, and
human--AI team performance. They should also report the relative
competence of the components and the allocation of decision authority
between them. Without such comparisons, improvement of human
performance under AI assistance establishes augmentation, but does not
establish genuine human--AI synergy.

\subsection{Chess as a Bridge Domain}
\label{sec:discussion-bridge}

The broader significance of the map is that chess provides a common
environment in which research traditions that are normally separated
can be compared. Human cognition contributes concepts such as expertise,
intuition, selective attention, bounded search, confidence, and
metacognition. Classical computer chess provides explicit models of
search and evaluation together with objective performance measures.
Neural and reinforcement-learning systems introduce learned
representations and policies. LLMs add language-mediated reasoning and
explanation, together with new questions about state tracking and
memorization. Hybrid systems allow these capabilities to be combined
and studied for complementarity.

These perspectives can often be applied to the same strategic decision.
A single chess position can be analyzed in terms of human expertise,
engine search and evaluation, neural representation, LLM explanation,
confidence, thinking time, and collaborative decision support. Few
domains combine formal structure, mature human expertise, superhuman
machine performance, and externally verifiable decision quality to the
same extent.

Standard chess nevertheless represents only one class of strategic
environment. Variants involving hidden information, modified rules,
unfamiliar initial states, or altered board structures can extend the
testbed toward uncertainty, information acquisition, and transfer.
They may therefore help distinguish capabilities tied to familiar chess
structures from more general forms of strategic reasoning.

The value of chess as a testbed consequently lies less in asking
whether machines can defeat humans---a question settled long ago---than
in using a controlled strategic environment to compare \emph{how}
different agents represent, evaluate, plan, explain, monitor, and
combine their decisions.

\subsection{Implications for Future Evaluation}
\label{sec:discussion-implications}

The mapping suggests that future evaluations should move beyond
single-dimensional measures of move quality or game outcome. Where
planning is claimed, final-action accuracy should be complemented by
evidence of look-ahead, plan formation, persistence, or revision.
Explanation-oriented systems should distinguish correctness and
faithfulness from fluency. Confidence should be evaluated for
calibration rather than accepted as linguistic self-report, and
assistance-oriented systems should be evaluated at the level of the
human--AI team.

Four priorities follow from these observations:

\begin{enumerate}[label=\textbf{P\arabic*.},leftmargin=*]
    \item \textbf{Explicit planning:}
    distinguish strong action selection from candidate generation,
    look-ahead, and longer-horizon plan formation;

    \item \textbf{Grounded and faithful explanation:}
    test whether explanations are strategically correct and supported
    by the decision process or external evidence;

    \item \textbf{Metacognitive calibration:}
    evaluate whether artificial agents recognize uncertainty, allocate
    reasoning effort appropriately, and revise unreliable conclusions;

 \item \textbf{Human--AI complementarity:}
    distinguish augmentation from genuine synergy by comparing
    human--AI team performance with the relevant constituent
    baselines, and investigate how complementarity changes with
    relative competence, task allocation, interaction design, and
    decision authority.
\end{enumerate}

These priorities are not intended as an exhaustive agenda for chess AI.
They identify research opportunities suggested by the sparse regions
of the systematic map.

\section{Threats to Validity}
\label{sec:threats}

As a systematic mapping study, this work is subject to threats related
to literature coverage, study selection, interpretive coding, and the
rapid evolution of the research area.

\paragraph{Search and selection.}
The interdisciplinary nature of chess research makes complete retrieval
difficult: relevant studies use heterogeneous terminology and appear
across computer science, AI, cognitive science, psychology, and related
fields. We mitigated this threat by combining five complementary
Scopus search families, Web of Science, and a curated Zotero collection.
Nevertheless, relevant studies may have been missed, particularly in
other databases or grey literature. Eligibility and study-family
consolidation also require judgment, especially when chess is only one
of several experimental domains or when the same contribution appears
as both a preprint and a published paper. The search queries, eligibility
criteria, and study-level decisions are provided in the replication
package to make these choices inspectable.

\paragraph{Coding and construct validity.}
The R1--R9 and E1--E10 frameworks require interpretive, multi-label
coding. To limit construct inflation, a code was assigned only when the
corresponding construct was directly studied, operationalized, or
evaluated; for example, strong move quality alone was not considered
evidence of search or strategic planning. All included study families
underwent a final verification pass in which overly broad assignments
were removed or refined. Independent duplicate coding was not performed
across the complete corpus, so coder subjectivity remains a limitation
and no inter-rater agreement statistic is reported. The framework
should moreover be understood as a mapping instrument, not as a claim
that strategic reasoning proceeds through nine discrete sequential
stages. Sparse or empty cells therefore indicate limited research under
our operational definitions, not the absence of the corresponding
capability.

The distinctions introduced in Sections \ref{sec:discussion-hybridization} and \ref{sec:discussion-synergy} were developed as post-hoc interpretive extensions of the mapping and were not applied as coding dimensions across the present corpus. Their empirical prevalence, discriminant value, and reproducibility as coding categories therefore remain to be evaluated in future work.

\paragraph{Temporal validity.}
The map focuses on the contemporary period beginning in 2023 in order
to capture the emergence of LLM-based approaches rather than reconstruct
the complete history of chess cognition and computer chess. This
excludes foundational earlier work from the quantitative analysis.
Publication cycles also differ across fields: LLM research is often
disseminated rapidly through preprints and conferences, whereas
cognitive and behavioral research may progress through slower journal
cycles. Because the field is evolving rapidly, the submitted map should
be interpreted as a reproducible snapshot defined by the documented
final search cutoff rather than as a permanently current census.

\paragraph{Interpretation of research gaps.}
Sparse regions of a systematic map identify underrepresented research
combinations, but do not by themselves establish scientific importance,
lack of capability, or lack of maturity. The gaps discussed in this
paper---explicit planning, grounded and faithful explanation,
metacognitive calibration, and human--AI complementarity---are therefore
interpreted as gaps in the \emph{distribution of recent research}.
Likewise, a densely populated cell indicates research attention, not
necessarily consensus or successful performance on that capability.

\section{Conclusion}
\label{sec:conclusion}

The recent chess literature provides a rich but uneven map of strategic reasoning. Research is strongest on assessment, evaluation, and action selection, while explicit planning, explanation, metacognition, and human--AI complementarity remain comparatively underdeveloped.

LLMs have expanded the field toward learned state representations, language-mediated reasoning, robustness, and explanation. Yet explanation-oriented work is concentrated in hybrid rather than purely language-model-based systems.

The mapping also revealed two future directions for refining the operational framework: hybrid systems are currently aggregated in a single category despite differing in where heterogeneous capabilities are combined and whether the coupling occurs at the moment of decision or across successive phases of learning and play; and the present evaluation framework does not distinguish improvements in human performance from genuine human–AI synergy, although the former alone cannot establish the latter. We therefore propose both distinctions as extensions of the framework, rather than treating them merely as findings about the literature.

We therefore identify four priorities for future chess-based strategic-reasoning research: explicit planning, grounded and faithful explanation, metacognitive calibration, and human–AI complementarity. Together, these directions shift the role of chess from a benchmark of playing strength toward a controlled environment for studying how human and artificial agents represent, plan, explain, monitor, and combine strategic decisions.

\section*{Statements and Declarations}

\paragraph{Funding.}
Remo Pareschi was partially supported by the project Ermete
(``Extended Reality for MaintenancE and Training Ecosystem''), funded
by the Italian Ministry of Enterprises and Made in Italy (MIMIT), 2023,
Grant number B39J25000400005.

\paragraph{Competing interests.}
The authors have no relevant financial or non-financial interests to disclose.

\paragraph{Data and code availability.}
The study protocol, search strategies, codebook, canonical study-family
dataset, and analysis script reproducing the principal quantitative
results are openly available in the replication package permanently
archived on Zenodo at
\url{https://doi.org/10.5281/zenodo.22695754}.

\paragraph{Use of generative AI.}
ChatGPT (OpenAI) and Claude (Anthropic) were used to support language editing, textual refinement,
and manuscript organization, and as a practical guide in preparing the
replication repository and its deposition on Zenodo. All AI-assisted
outputs were reviewed and revised by the authors, who take full
responsibility for the final manuscript and accompanying research
materials.

\printbibliography

\end{document}